\documentclass{article}
\usepackage{spconf,amsmath,graphicx}
\usepackage{times}
\usepackage{latexsym}

\usepackage[T1]{fontenc}
\usepackage[utf8]{inputenc}
\usepackage{microtype}
\usepackage{inconsolata}
\usepackage{amsmath}
\usepackage{subfigure}

\usepackage{textcomp}
\usepackage{stfloats}
\usepackage{url}
\usepackage{verbatim}
\usepackage{graphicx}
\usepackage{cite}
\usepackage{booktabs}
\usepackage{longtable}
\usepackage{threeparttable}
\usepackage{pifont}
\usepackage{arydshln}
\usepackage{svg}
\usepackage{multirow}
\usepackage{amsfonts,amssymb}

\title{MSG-BART: Multi-granularity Scene Graph-Enhanced Encoder-Decoder Language Model for Video-grounded Dialogue Generation}
\name{HongchengLiu, Zhe Chen,  Hui Li,  Pingjie Wang, Yanfeng Wang, Yu Wang*}
\address{Cooperative Medianet Innovation Center, Shanghai Jiao Tong University}

\begin{document}
\ninept
\maketitle

\begin{abstract}
Generating dialogue grounded in videos requires a high level of understanding and reasoning about the visual scenes in the videos. However, existing large visual-language models are not effective due to their latent features and decoder-only structure, especially with respect to spatio-temporal relationship reasoning. In this paper, we propose a novel approach named MSG-BART, which enhances the integration of video information by incorporating a multi-granularity spatio-temporal scene graph into an encoder-decoder pre-trained language model. Specifically, we integrate the global and local scene graph into the encoder and decoder, respectively, to improve both overall perception and target reasoning capability. To further improve the information selection capability, we propose a multi-pointer network to facilitate selection between text and video. Extensive experiments are conducted on three video-grounded dialogue benchmarks, which show the significant superiority of the proposed MSG-BART compared to a range of state-of-the-art approaches.
\end{abstract}
\section{Introduction}

Large visual-language models (VLMs) with CNN-based or ViT-based (CNN/ViT) features \cite{llava} have recently shown remarkable capabilities in many vision tasks, including video-grounded dialogue (VGD) systems. However, VGD systems need to consider temporal information from dynamic, unconstrained visual scenes, which is closer to real life and more challenging \cite{pasunuru-bansal-2018-game}. Despite they make great progress in VGD, large VLMs in VGD still lack efficient temporal sensitivity and the ability to incorporate the entire spatio-temporal dimension into video representation \cite{videollama}. Therefore, the main challenge for existing VGD methods is how to effectively utilize video information, particularly spatio-temporal information.
\begin{figure}[t]
\centering
\includegraphics[width=0.8\linewidth]{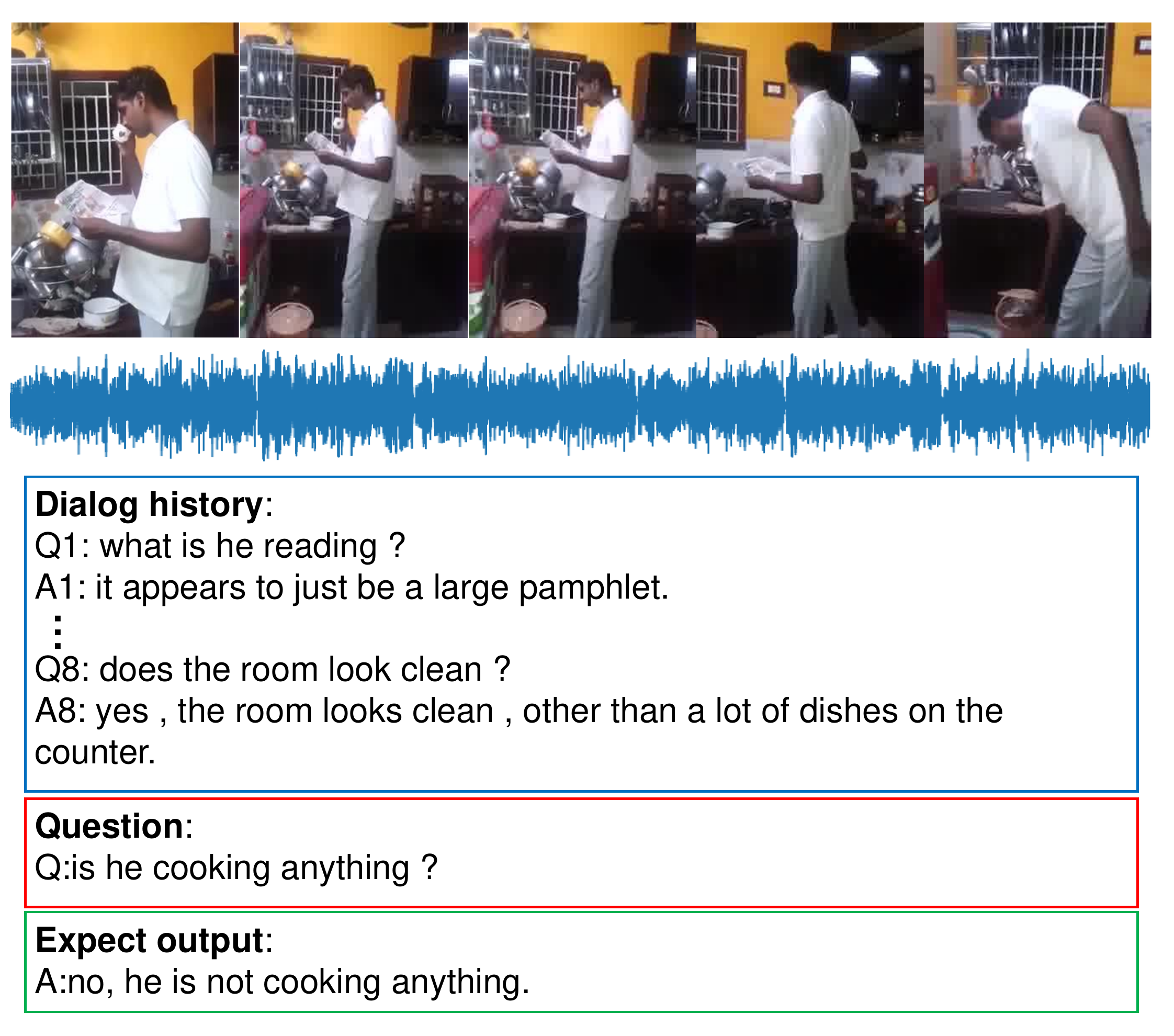}
\caption{The demonstration of the VGD task. 
}
\label{fig:avsd}
\setlength{\belowcaptionskip}{-5cm} 
\end{figure}
However, the CNN/ViT-based features are inefficient in spatio-temporal reasoning because they cannot capture and understand the essential entities, relationships, and temporal dynamics involved \cite{kim2021structured}. On the other hand, scene graphs are widely used in event detection because they can provide efficient semantic information that can represent visual scenes \cite{khademi2020multimodal}. Given the importance of spatio-temporal perception in VGD, we need to include scene graphs to establish an integrated understanding capability.


One way to improve the use of scene graphs is by implementing the encoder-decoder (ED) architecture instead of the decoder-only architecture for semantic extraction. Fu et al \cite{fu2023decoder} suggest that the decoder-only structure is insufficient for temporal awareness as it focuses less on the source input information. In addition, recent researches \cite{kriegeskorte2019interpreting,videollm} demonstrate that the encoder can still outperform the decoder in comprehensive predictions, which is necessary for scene graph processing. From this perspective, it is recommended to use both the encoder and decoder of the language models for semantic extraction and integration with scene graphs.
Furthermore, to fully leverage the potential of the scene graph in the ED architecture, the encoder and the decoder of the language models should be used for the global perception and detail-sensitive features, respectively \cite{liu2021kg}. In other words, the ED framework needs to combine features of different granularity. 
To this end, \cite{wang2020ord} recognized the importance of graph features at different granularities but only injects the entire graph into the encoder and ignores the impact of multi-granularity graph information. Therefore, in this paper, we propose a hierarchical-based method to obtain multi-granularity scene graph information.

Another way to improve the use of scene graphs is to leverage a more powerful selection mechanism for multi-modality information combinations. Many existing VLMs rely on different attention scores between tokens to do selection \cite{videogpt}, but they cannot effectively select information from different modalities because they do not treat the same modality as a unified entity. Based on this consideration, \cite{le2020multimodal} adopt the Pointer-Generator Network to combine the generation result and text information as the final responses. Although this approach effectively improves the utilization of text, it combines information via a single pointer and ignores the need to compare various modalities to determine the final result when these are all suitable for generation in combination. For this reason, we calculate the different selection scores separately for different modalities.

In this paper, we propose a novel multi-granularity scene graph model, MSG-BART, which enhances the BART\footnote{Pre-trained language model with the encoder-decoder Transformer structure.} model \cite{lewis2020bart} with fine-grained cross-modal structured knowledge. The main novelty of MSG-BART is two-fold.
First, we effectively incorporate the global spatio-temporal scene graphs in the encoder for the comprehensive perception of the entire video and local graphs in the decoder for detailed reasoning of the video segment.
Second, we propose a multi-pointer network to enhance the video-text information selection capability. 
To achieve this, we design a Graph-and-Video Processing (GVP) module including a GVP encoder and a GVP decoder for semantic extraction and fusion of text, vision, and audio. The encoder is mainly responsible for the overall perception of the video, and the decoder is mainly responsible for the targeted addition of detailed information. Additionally, we propose a multi-pointer network for information selection, which calculates multiple-pointer values to achieve the most appropriate combination between text and video.

To evaluate the performance of the proposed MSG-BART method, we compare it with a range of state-of-the-art methods on three VGD benchmarks: DSTC8-AVSD, DSTC10-AVSD, and NExT-OE datasets. The results show that the performance of MSG-BART substantially outperforms existing methods in most benchmarks, validating its superiority and effectiveness.

\section{Method}

\subsection{Task Formulation}
\begin{figure}[t]
\centering
\includegraphics[width=0.75\linewidth]{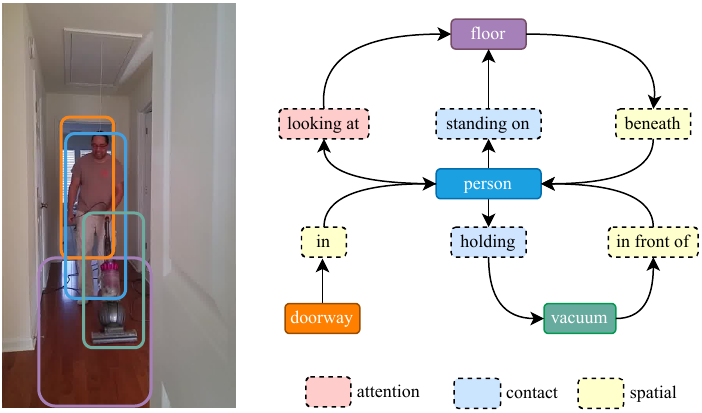}
\caption{The demonstration of the scene graph. 
}
\label{fig:sce}
\end{figure}

Video-grounded dialogue task aims to generate appropriate and fluent responses $Y=\{y_1,y_2,...,y_t \}$ containing $t$ words, in the condition of the quadruple $(V, A, D, Q)$ as inputs where $D$ is dialogue history, $V$ is video, $A$ is audio and $Q$ is question. Here the dialogue history consists of multi-turn question-answer pairs. For video information utilization, we extract multi-granularity visual features and audio features for different time durations by ActionCLIP \cite{wang2021actionclip} and Wav2CLIP \cite{wav2clip}. Additionally, We use STTran \cite{sttran} to generate the spatio-temporal scene graph $\mathcal{G}$, and can be seen in Fig.~\ref{fig:sce}. It can be formulated as $\mathcal{G}=\{(x_i,r_{ij},x_j)|x_i,x_j\in\nu,r_{ij}\in \varepsilon\}$, where $\nu=\{x_1,...,x_N\}$ denotes the set of $N$ object nodes and $\varepsilon$ represents the set of $M$ relational edges between  source node $x_i$ and target node $x_j$. 

\subsection{Overview}   

\begin{figure*}[h]
\centering
\includegraphics[width=\linewidth]{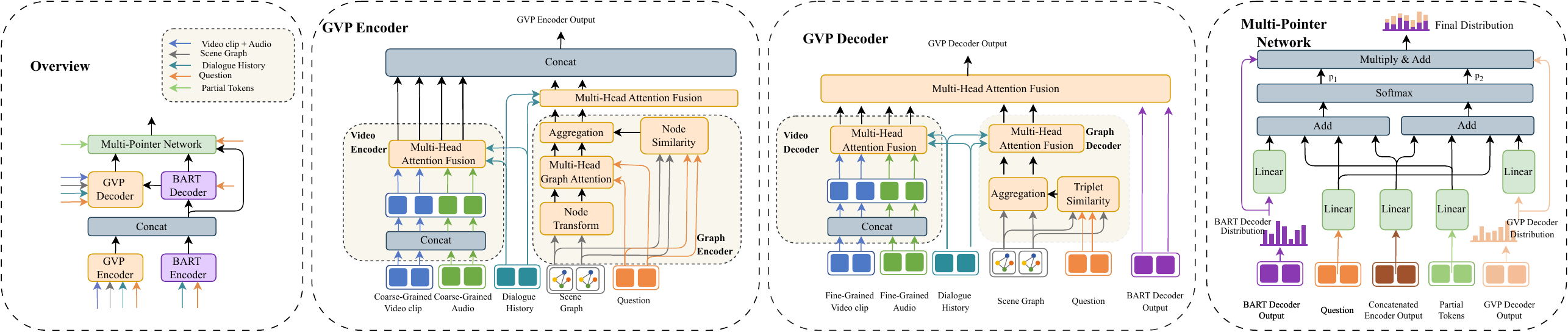}

\caption{Illustration of the MSG-BART which consists of five parts: the BART encoder, BART decoder, \textbf{GVP encoder}, \textbf{GVP decoder}, and \textbf{Multi-Pointer Network}. 
}
\label{fig:model}
\end{figure*}

The overall architecture of the MSG-BART model is shown in Fig.~\ref{fig:model}, which consists of the encoder and decoder of BART \cite{lewis2020bart}, the graph-video processing modules, and a multi-pointer network. To be specific, we use the BART encoder to model the dialogue history and the BART decoder to generate responses according to the question. To enhance the ability to incorporate multimodal information, we propose a GVP module, in which the encoder extracts effective representations for videos by inferring coarse-grained audio-visual features and global scene graphs via node form, and the decoder injects fine-grained audio-visual features and local scene graphs in the form of triplets into the output of the BART decoder. The final multi-pointer network computes p-values for the output of the GVP decoder and the BART decoder respectively to selectively combine them. In the following, the details of each component will be described. 

\subsection{GVP Encoder}
We devise a GVP encoder to enhance the BART encoder to obtain a rich representation of the video. The GVP encoder contains the video encoder and the graph encoder, which are responsible for coarse-grained audio-visual feature inference and global spatio-temporal scene graph inference in the form of graph nodes, respectively.

\subsubsection{Video Encoder}
To make better use of the audio-visual features, we propose video co-attention to enhance the semantic information. Specifically, coarse-grained video clip features and audio features are concatenated into the video features $V_c$, and then the video features are aligned with the dialogue history in multi-head attention to obtain the video encoder output $Y_{\rm v\_en}$ as
\begin{equation}
    Y_{\rm v\_en} = \operatorname{FFN}(V_{\rm c} +\operatorname{MHA}(V_{\rm c},D,D)),
\label{eq:Y_{v_en}}
\end{equation}
where $\operatorname{FFN}$ and $\operatorname{MHA}$ denote the feed-forward network and multi-head attention, and the input of $\operatorname{MHA}(\cdot,\cdot,\cdot)$ represent the query, key and value respectively.

\subsubsection{Graph Encoder}
\textbf{Node Transform \& Node Similarity}
The scene graph contains entities in a video clip and the relations between them. However, many of these entities and relations are not relevant for answering questions. To address this issue, we propose a similarity-based node filtering method. This method selects the most relevant nodes for the questions, reducing redundancy.

Similar to DialoKG \cite{dialokg}, we first convert the relational edges of the spatio-temporal scene graph into the node level, namely, relational node (RN) scene graph. 
The transformed scene graph is formulated as $\mathcal{G}_{\rm R}=\{ (x_i,e_{ij},x_j)|x_i,x_j\in \nu+\varepsilon,e_{ij} \in \{0,1\} \}$, where $\nu+\varepsilon$ denotes the set of $N$ object nodes and $M$ relational nodes, and $e_{ij}$ represents edge used to indicate whether there is a directed link between node $x_i$ and $x_j$. With the converted graph, we then calculate the node similarity between each node and the question to select the most important node for node aggregation. The node similarity $S_n$ is computed with the degree matrix $M_{\rm D}$ and adjacency matrix $M_{\rm A}$ as
\begin{equation}
    S_{ n}=M_{\rm D}^{-1}\ast (M_{\rm A}+I) \ast S_{\rm c},
\end{equation}
where $I$ denotes the identity matrix, and $S_{\rm c}= \operatorname{Cosine}(X,Q)$ denotes the cosine similarity scores of node features $X$ and the question.

\textbf{ Node Updating}
To fully exploit the information in scene graphs, we propose a question-aware attention mechanism to obtain the spatio-temporal scene graph representation. Specifically, we utilize a multi-head graph attention module for node updating and graph aggregation for information aggregation.

For node features $X=\{x_1, \ldots, x_{M+N}\}$ in the RN scene graph, multi-head attention is performed between the source and target nodes. The attention weight $\alpha_{h,ij}$ estimates the degree of the correlation of source node $x_i$ and target node $x_j$ for question in $h$th attention head, which is formulated as
\begin{equation}
    \alpha_{h,ij} = \frac{\operatorname{exp}(\operatorname{\sigma} (W_1 Q + W_2 x_i + W_3 x_j ))}{{\textstyle \sum\limits_{x_k \in X_j}}\operatorname{exp}( \sigma (W_1 Q + W_2 x_k + W_3 x_j ))},
\end{equation}
where $X_j$ denotes the set of source nodes with a directed link to the same target node $x_j$, $W_1$, $W_2$, $W_3$ are learned parameters, and $\operatorname{\sigma}$ denotes the sigmoid function. The node feature $x_j$ are updated to $x'_j$ with attention weight $\alpha_{h,ij}$ as
\begin{equation}
    x'_{j}=\operatorname{FFN}(\operatorname{Concat}({\textstyle \sum\limits_{i \in N_i}{\alpha_{h,ij}x_i} })|_{h=0}^{H}),
\end{equation}
where $H$ represents the number of heads.

\textbf{Graph Aggregation \& Attention Fusion}
The node features $X'=\{x'_1,\ldots,x'_{M+N}\}$ contains rich semantic information after graph attention, and the importance scores $S_n=\{s_1,...,s_{M+N}\}$ for all the nodes to answer the question are obtained by node similarity. We select the node feature of the most important one according to the node similarity as the RN scene graph representation $G_r$ of each image to aggregate information, which can be formulated as
\begin{equation}
    G_r=\{x'_{j}|j=\operatorname{index}(\operatorname{max}(S_{ n})),x'_j\in X'\}
\end{equation}

The graph encoder output 
is enhanced by aligning the graph representation $G_r$ with the dialogue history, and the calculation is similar to Eq. \ref{eq:Y_{v_en}}.

\subsection{GVP Decoder}

The GVP decoder injects relevant multimodal information into the output of the BART decoder. To enrich the model output with more fine-grained features, we enhance the BART decoder with fine-grained video and local spatio-temporal scene graph triplets through a multi-head attention fusion module.
The GVP decoder consists of a video and a graph decoder. The architecture of the video decoder is the same as that of the video encoder.

\textbf{Triplet Similarity \& Aggregation}
The final output is more sensitive to the input of the decoder as it is closer to the final process, so we adopt the scene graph of triplet node form to inject the most appropriate local information. 
Specifically, the transformed scene graph of triplet nodes are fully connected and the node set is $N_{\rm T}=\{n_{\rm T}^k|n_{\rm T}^k=[x_i,r_{ij},x_j],(x_i,r_{ij},x_j)\in\mathcal{G},k\in\{1,...,M\}\}$. In this case, the triplet similarity between triplet nodes and the question is derived from the cosine similarity of them as $S_t=\operatorname{Cosine}(N_{\rm T}, Q)$.

When aggregation, the triplet node feature with the highest triplet similarity is considered to be the most significant one and selected as the graph representation $G_t$, which is formulated as
\begin{equation}
    G_t=\{n_{\rm T}^i|i=\operatorname{index}(\operatorname{max}(S_t)),n_{\rm T}^i\in N_{\rm T}\}.
\end{equation}

\textbf{Attention Fusion}
To better utilize the fine-grained video features and local scene graphs, we enhance semantic representations by aligning them with the dialogue history, in which the output of graph decoder $Y_{\rm g\_de}$ is derived similarly to Eq.~\ref{eq:Y_{v_en}}.

Finally, considering the video decoder output as $Y_{\rm v\_de}$, the graph decoder output as $Y_{\rm g\_de}$, and the BART decoder output as $Y_{\rm B\_de}$, $Y_{\rm v\_de}$ and $Y_{\rm g\_de}$ are first concatenated, and the output of the GVP decoder $Y_{\rm G\_de}$ can be derived as
\begin{equation}
    \begin{aligned}
        Y_{\rm G\_de}=&\operatorname{FFN}(Y_{\rm B\_de}+\operatorname{MHA}(Y_{\rm B\_de}, \\ 
        & [Y_{\rm v\_de},Y_{\rm g\_de}],[Y_{\rm v\_de},Y_{\rm g\_de}])).
    \end{aligned}
\end{equation}

\subsection{Multi-Pointer Network}

The outputs of the BART decoder and the GVP decoder are biased differently. The BART decoder is more biased toward semantic extraction of the dialogue history and the overall perceptual representation of the video, while the GVP decoder is more biased toward fine-grained semantic information of the video itself. Therefore, we first calculate the correlation between the outputs of different modules and the final result to obtain $p'_i$ for information selection as
\begin{equation}
    \left\{
    \begin{aligned}
        p_1^{\prime} &=W_4 Y_{\rm B\_de}+W_5 Y_{<t}+W_6 Y_{\rm en}+W_7 Q \\
        p_2^{\prime} &=W_8 Y_{\rm G\_de}+W_5 Y_{<t}+W_6 Y_{\rm en}+W_7 Q,
    \end{aligned}
    \right.
\end{equation}
where $W_4, W_5, W_6, W_7, W_8$ are the learnable weights, $Y_{<t}$ denotes the partial tokens, $Y_{\rm en}$ denotes the encoder state, $Y_{\rm B\_de}$ denotes the BART decoder state, $Y_{\rm G\_de}$ denotes the GVP decoder state.

$p'$ are then concatenated and passed through $\operatorname{Softmax}$ to obtain p-values $p_i$ as
\begin{equation}
    [p_1,p_2]=\operatorname{Softmax}([p_1',p_2']).
\end{equation}

The final output $\hat{Y}_t$ is calculated as:
\begin{equation}
    \hat{Y}_t = p_1 \ast Y_{\rm G\_de} + p_2 \ast Y_{\rm B\_de}
\end{equation}
where $Y_{\rm G\_de}$ and $Y_{\rm B\_de}$ denote the GVP decoder output distribution and BART decoder output distribution, respectively.

\subsection{Loss Function}
The loss function is a standard cross-entropy loss:
\begin{equation}
    \mathcal{L}=-\sum_{t=1}^{T} \log P\left(y_t \mid V, A, \mathcal{G}, D, Q, Y_{<t}\right),
\end{equation}

\section{Experiments}

\begin{table}[t]
\centering
\setlength\tabcolsep{10pt}
\resizebox{1\linewidth}{!}
{
\begin{tabular}{lcccc}
\bottomrule[1pt]
\textbf{Models}  &\bf BLEU-4 &\bf METEOR &\bf ROUGH-L &\bf CIDEr \\ \hline
\multicolumn{5}{c}{\it DSTC8-AVSD official test set}    \\ \hline
DMN~\cite{xie2020audio}&0.270 &0.208 &0.482 &0.714 \\
VideoGPT (DSTC8 bset)~\cite{videogpt}      & 0.387       & 0.249       & 0.544        & 1.022      \\
SCGA~\cite{kim2021structured} &0.377 &0.269 &0.555 &1.024 \\
MED-CAT~\cite{med-cat}&0.376 &0.247 &0.547 &0.982 \\
MSG-BART (Ours)   & \textbf{0.403}   & \textbf{0.270}   & \textbf{0.569}    & \textbf{1.105}  \\ \hline
\multicolumn{5}{c}{\it DSTC10-AVSD official test set }  \\ \hline
Ext-AV-trans~\cite{av-trans}& 0.371       & 0.245       & 0.535        & 0.869      \\
UniVL-obj (DSTC10 best)~\cite{med-cat}& 0.372       & 0.243       & 0.530        & 0.912      \\
TSF(ensemble)~\cite{timesformer-gpt}   &0.385& 0.247 &0.539 &0.957      \\
DialogMCF~\cite{DialogMCF} &0.369 &0.249 &0.536 &0.912      \\
MSG-BART (Ours)    & \textbf{0.390}   & \textbf{0.268}   & \textbf{0.556}    & \textbf{1.008}  \\ \bottomrule[1pt]
\end{tabular}
}
\caption{Evaluation results of our model compared with baseline approaches on DSTC8-AVSD and DSTC10-AVSD official test sets.}
\label{tab:table-best}

\end{table}

\subsection{Dataset and Metrics}
 \textbf{The AVSD datasets} were generated by expanding the Charades dataset with Q$\&$A, where the dialogue contains open-ended question-answer pairs regarding the scenes in the video and the length of pairs is mostly more than 15 words. To evaluate the accuracy and diversity of the methods, we used BLEU, METEOR, ROUGE-L and CIDEr as the evaluation metrics. 
\textbf{The NExT-OE dataset} is constructed based on YFCC-100M, where the questions are generated by casualty, temporary reasoning, and descriptive ability.  Different from the DSTC-AVSD, the answers are mostly shorter than 4 words and the evaluation metric is the WUPS score to evaluate the semantic similarity between the answer and ground truth.

\subsection{Experimental Setup}
\label{sec:appendix-setup}
We initialized our model using the weights of the BART-base\footnote{\url{https://huggingface.co/facebook/bart-base}} model. In the training phase, we used 4 heads in the multi-head attention model, the hidden size was 768 and the batch size was 32. We adopted an AdamW optimizer with a learning rate for fine-tuning BART of 6.25e-5 and a learning rate of 6.25e-4 for training the GVP module. During the decoding phase, we used the beam search algorithm with a beam size of 6 and a penalty factor of 0.6.

\begin{table}[t]
\tiny
\centering
\resizebox{0.8\linewidth}{!}
{
\begin{tabular}{lc}
\bottomrule[1pt]
 \textbf{Models}           & \textbf{WUPS} \\ \hline
UATT    \cite{xue2017unifying}     & 24.25  \\
HGA      \cite{jiang2020reasoning}    & 25.18  \\
ClipBERT  \cite{lei2021less}   & 24.17  \\
KcGA \cite{jin2023knowledge}(with external knowledge)     & \textbf{28.20}  \\ \hline
MSG-BART(ours)             & 27.45  \\
MSG-BART(ours)(data augmentation) & \textbf{29.44}  \\ \bottomrule[1pt]
\end{tabular}}
\caption{Evaluation results of our model compared with baseline approaches on the  NExT-OE dataset.}
\label{tab:table-nextoe}
\end{table}

\subsection{Main Experimental Results}
The main experimental results are shown in Table~\ref{tab:table-best} and Table~\ref{tab:table-nextoe}. It shows that the proposed MSG-BART method can outperform the baseline methods on most metrics by a large margin on four datasets. 
For AVSD dataset, it can be seen that on the BLEU4 and CIDEr metrics, which reflect the key information generation and language fluency, MSG-BART yields the most performance gains. 
Compared to VideoGPT (the best performance in the decoder-only model), MSG-BART yields significant performance improvement, which validates the advantage of the proposed GVP module for multimodal information processing and understanding. Additionally, MSG-BART also gives superior performance than the Ext-AV-trans methods, which shows the effectiveness of the proposed multi-granularity scene graph in capturing the cross-modal information.
For NExT-OE dataset,  the result outperforms most baselines in the condition that lots of noise introduced by scene graph generator is not adaptive to the NExT-OE videos. To confirm our configuration, we conduct dialogue history with the questions that are irrelevant to answers, and the MSG-BART after data augmentation surpasses all the methods.


\subsection{Ablation Study}

To explore the contribution of each proposed component, we conduct ablation experiments, and the results are shown in Table~\ref{tab:table-ablation-model}. 
In terms of information incorporation, there is a significant degradation in performance after removing each proposed component, which validates their importance for information extraction from scene graphs. Specifically, removing the similarity calculation module, especially the triplet similarity degrades the performance for all cases. This is caused by the question-irrelevant noise introduced in the graph features, especially the fine-grained graph features without similarity processing. Also, the experimental results suggest that both graph encoder and graph decoder are important to capture the spatio-temporal information and also the action behavior information embedded in the scene graph, especially the graph encoder. 

In terms of information selection, the results listed in the second block of Table~\ref{tab:table-ablation-model} show that the multi-pointer network can yield performance gains. In detail, we formulate $p_1=p_2=0.5$ to remove the pointer and replace $p_2$ with $1-p_1$ for the single pointer setting. It can be seen that using a single pointer worsens the performance compared to removing the pointer network completely, which demonstrates that a single pointer cannot select the most appropriate information based on the compatibility of textual information in cases where video information is more appropriate, and confirms the advantage of our multi-pointer network in information selection and fusion. 
\begin{table}[tb]
\setlength\tabcolsep{3.5pt}
\centering
\resizebox{0.9\linewidth}{!}
{
\begin{tabular}{lcccc}
\bottomrule[1pt]
\textbf{Method} & \bf BLEU-4 &\bf METEOR &\bf ROUGH-L &\bf CIDEr \\ \hline
Full      & \textbf{0.390} & \textbf{0.268} & \textbf{0.556} & \textbf{1.008} \\
- GAT     & 0.381     & 0.263     & 0.545     & 0.982     \\
- Node-similarity  & 0.380     & 0.263     & 0.546     & 0.988     \\
- Triplet-similarity  & 0.378     & 0.262     & 0.545     & 0.977     \\ 
- Graph-encoder & 0.380     & 0.264     & 0.547     & 0.989     \\
- Graph-decoder & 0.387     & 0.267     & 0.552     & 0.995     \\ \hline
- Pointer      & 0.381     & 0.266     & 0.551     & 0.996     \\
Single pointer & 0.378     & 0.263     & 0.549     & 0.989     \\\bottomrule[1pt]
\end{tabular}}
\caption{Evaluation results of module ablation experiments on DSTC10-AVSD official test set.}
\label{tab:table-ablation-model}
\end{table}
\section{Conclusions}
To improve the generation of video-grounded dialogue, we have proposed a novel model named MSG-BART, in which a GVP module is devised to enhance the incorporation of multimodal information into the encoder and decoder, especially multi-granularity spatio-temporal scene graphs. Furthermore, we propose a multi-pointer network for active information selection. Extensive experiments conducted on three VGD benchmarks consistently verify the superiority of the proposed methods compared to a range of state-of-the-art methods. 

\bibliographystyle{IEEEbib}
\bibliography{custom.bib}

\begin{thebibliography}{10}

\bibitem{llava}
Haotian Liu, Chunyuan Li, Qingyang Wu, and Yong~Jae Lee,
\newblock ``Visual instruction tuning,''
\newblock {\em arXiv preprint arXiv:2304.08485}, 2023.

\bibitem{pasunuru-bansal-2018-game}
Ramakanth Pasunuru and Mohit Bansal,
\newblock ``Game-based video-context dialogue,''
\newblock in {\em Proceedings of the 2018 Conference on Empirical Methods in
  Natural Language Processing}, Brussels, Belgium, Oct.-Nov. 2018, pp.
  125--136.

\bibitem{videollama}
Hang Zhang, Xin Li, and Lidong Bing,
\newblock ``Video-llama: An instruction-tuned audio-visual language model for
  video understanding,''
\newblock {\em arXiv preprint arXiv:2306.02858}, 2023.

\bibitem{kim2021structured}
Junyeong Kim, Sunjae Yoon, Dahyun Kim, and Chang~D Yoo,
\newblock ``Structured co-reference graph attention for video-grounded
  dialogue,''
\newblock in {\em Proceedings of the AAAI Conference on Artificial
  Intelligence}, 2021, vol.~35, pp. 1789--1797.

\bibitem{khademi2020multimodal}
Mahmoud Khademi,
\newblock ``Multimodal neural graph memory networks for visual question
  answering,''
\newblock in {\em Proceedings of the 58th Annual Meeting of the Association for
  Computational Linguistics}, 2020, pp. 7177--7188.

\bibitem{fu2023decoder}
Zihao Fu, Wai Lam, Qian Yu, Anthony Man-Cho So, Shengding Hu, Zhiyuan Liu, and
  Nigel Collier,
\newblock ``Decoder-only or encoder-decoder? interpreting language model as a
  regularized encoder-decoder,''
\newblock {\em arXiv preprint arXiv:2304.04052}, 2023.

\bibitem{kriegeskorte2019interpreting}
Nikolaus Kriegeskorte and Pamela~K Douglas,
\newblock ``Interpreting encoding and decoding models,''
\newblock {\em Current opinion in neurobiology}, vol. 55, pp. 167--179, 2019.

\bibitem{videollm}
Guo Chen, Yin-Dong Zheng, Jiahao Wang, Jilan Xu, Yifei Huang, Junting Pan,
  Yi~Wang, Yali Wang, Yu~Qiao, Tong Lu, et~al.,
\newblock ``Videollm: Modeling video sequence with large language models,''
\newblock {\em arXiv preprint arXiv:2305.13292}, 2023.

\bibitem{liu2021kg}
Ye~Liu, Yao Wan, Lifang He, Hao Peng, and S~Yu Philip,
\newblock ``Kg-bart: Knowledge graph-augmented bart for generative commonsense
  reasoning,''
\newblock in {\em Proceedings of the AAAI Conference on Artificial
  Intelligence}, 2021, vol.~35, pp. 6418--6425.

\bibitem{wang2020ord}
Ziwei Wang, Zi~Huang, Yadan Luo, and Huimin Lu,
\newblock ``Ord: Object relationship discovery for visual dialogue
  generation,''
\newblock {\em arXiv preprint arXiv:2006.08322}, 2020.

\bibitem{videogpt}
Zekang Li, Zongjia Li, Jinchao Zhang, Yang Feng, and Jie Zhou,
\newblock ``Bridging text and video: A universal multimodal transformer for
  audio-visual scene-aware dialog,''
\newblock {\em IEEE/ACM Transactions on Audio, Speech, and Language
  Processing}, vol. 29, pp. 2476--2483, 2021.

\bibitem{le2020multimodal}
Hung Le and Nancy~F Chen,
\newblock ``Multimodal transformer with pointer network for the {DSTC8} {AVSD}
  challenge,''
\newblock {\em ArXiv preprint}, vol. abs/2002.10695, 2020.

\bibitem{lewis2020bart}
Mike Lewis, Yinhan Liu, Naman Goyal, Marjan Ghazvininejad, Abdelrahman Mohamed,
  Omer Levy, Veselin Stoyanov, and Luke Zettlemoyer,
\newblock ``{BART}: Denoising sequence-to-sequence pre-training for natural
  language generation, translation, and comprehension,''
\newblock in {\em Proceedings of the 58th Annual Meeting of the Association for
  Computational Linguistics}, 2020, pp. 7871--7880.

\bibitem{wang2021actionclip}
Mengmeng Wang, Jiazheng Xing, and Yong Liu,
\newblock ``{ActionCLIP}: A new paradigm for video action recognition,''
\newblock {\em ArXiv preprint}, vol. abs/2109.08472, 2021.

\bibitem{wav2clip}
Ho-Hsiang Wu, Prem Seetharaman, Kundan Kumar, and Juan~Pablo Bello,
\newblock ``{Wav2CLIP}: Learning robust audio representations from {CLIP},''
\newblock in {\em ICASSP 2022-2022 IEEE International Conference on Acoustics,
  Speech and Signal Processing}. IEEE, 2022, pp. 4563--4567.

\bibitem{sttran}
Yuren Cong, Wentong Liao, Hanno Ackermann, Bodo Rosenhahn, and Michael~Ying
  Yang,
\newblock ``Spatial-temporal transformer for dynamic scene graph generation,''
\newblock in {\em Proceedings of the IEEE/CVF International Conference on
  Computer Vision}, 2021, pp. 16372--16382.

\bibitem{dialokg}
Md~Rashad Al~Hasan Rony, Ricardo Usbeck, and Jens Lehmann,
\newblock ``{D}ialo{KG}: Knowledge-structure aware task-oriented dialogue
  generation,''
\newblock in {\em Findings of the Association for Computational Linguistics:
  NAACL 2022}, Seattle, United States, 2022, pp. 2557--2571.

\bibitem{xie2020audio}
Huiyuan Xie and Ignacio Iacobacci,
\newblock ``Audio visual scene-aware dialog system using dynamic memory
  networks,''
\newblock {\em DSTC8 at AAAI2020 workshop}, 2020.

\bibitem{med-cat}
Xin Huang, Hui~Li Tan, Mei~Chee Leong, Ying Sun, Liyuan Li, Ridong Jiang, and
  Jung-jae Kim,
\newblock ``Investigation on transformer-based multi-modal fusion for
  audio-visual scene-aware dialog,''
\newblock in {\em Proceedings of DSTC10 Workshop at AAAI-2022}, 2022.

\bibitem{av-trans}
Ankit Shah, Shijie Geng, Peng Gao, Anoop Cherian, Takaaki Hori, Tim~K Marks,
  Jonathan Le~Roux, and Chiori Hori,
\newblock ``Audio-visual scene-aware dialog and reasoning using audio-visual
  transformers with joint student-teacher learning,''
\newblock in {\em ICASSP 2022-2022 IEEE International Conference on Acoustics,
  Speech and Signal Processing}, 2022, pp. 7732--7736.

\bibitem{timesformer-gpt}
Yoshihiro Yamazaki, Shota Orihashi, Ryo Masumura, Mihiro Uchida, and Akihiko
  Takashima,
\newblock ``Audio visual scene-aware dialog generation with transformer-based
  video representations,''
\newblock {\em ArXiv preprint}, vol. abs/2202.09979, 2022.

\bibitem{DialogMCF}
Zhe Chen, Hongcheng Liu, and Yu~Wang,
\newblock ``Dialogmcf: Multimodal context flow for audio visual scene-aware
  dialog,''
\newblock {\em IEEE/ACM Transactions on Audio, Speech, and Language
  Processing}, pp. 1--13, 2023.

\bibitem{xue2017unifying}
Hongyang Xue, Zhou Zhao, and Deng Cai,
\newblock ``Unifying the video and question attentions for open-ended video
  question answering,''
\newblock {\em IEEE Transactions on Image Processing}, vol. 26, no. 12, pp.
  5656--5666, 2017.

\bibitem{jiang2020reasoning}
Pin Jiang and Yahong Han,
\newblock ``Reasoning with heterogeneous graph alignment for video question
  answering,''
\newblock in {\em Proceedings of the AAAI Conference on Artificial
  Intelligence}, 2020, vol.~34, pp. 11109--11116.

\bibitem{lei2021less}
Jie Lei, Linjie Li, Luowei Zhou, Zhe Gan, Tamara~L Berg, Mohit Bansal, and
  Jingjing Liu,
\newblock ``Less is more: Clipbert for video-and-language learning via sparse
  sampling,''
\newblock in {\em Proceedings of the IEEE/CVF conference on computer vision and
  pattern recognition}, 2021, pp. 7331--7341.

\bibitem{jin2023knowledge}
Yao Jin, Guocheng Niu, Xinyan Xiao, Jian Zhang, Xi~Peng, and Jun Yu,
\newblock ``Knowledge-constrained answer generation for open-ended video
  question answering,''
\newblock in {\em Proceedings of the AAAI Conference on Artificial
  Intelligence}, 2023, vol.~37, pp. 8141--8149.

\end{thebibliography}
\end{document}